\documentclass[twoside,11pt]{article}

\usepackage{jmlr2e}

\usepackage[english]{babel}
\usepackage{booktabs}   
\usepackage{tikz}
\usepackage{pifont}
\newcommand{\cmark}{\ding{51}}
\newcommand{\xmark}{\ding{55}}

\usepackage[utf8]{inputenc} 
\usepackage[T1]{fontenc}    
\usepackage{hyperref}       
\usepackage{url}            
\usepackage{amsfonts}       
\usepackage{microtype}      
\usepackage{graphicx} 

\usepackage{listings}
\usepackage{xcolor}

\definecolor{codegreen}{rgb}{0,0.6,0}
\definecolor{codegray}{rgb}{0.5,0.5,0.5}
\definecolor{codepurple}{rgb}{0.58,0,0.82}
\definecolor{backcolour}{gray}{0.9}

\lstdefinestyle{mystyle}{
    backgroundcolor=\color{backcolour},   
    commentstyle=\color{codegreen},
    keywordstyle=\color{magenta},
    numberstyle=\tiny\color{codegray},
    stringstyle=\color{codepurple},
    basicstyle=\ttfamily\footnotesize,
    breakatwhitespace=false,         
    breaklines=true,                 
    captionpos=b,                    
    keepspaces=true,                 
    numbers=left,                    
    numbersep=5pt,                  
    showspaces=false,                
    showstringspaces=false,
    showtabs=false,                  
    tabsize=2
}
\makeatletter

\newcommand{\crl}{\texttt{ChainerRL}}
\newcommand{\crlv}{\texttt{ChainerRL Visualizer}}

\usepackage{pifont}

\usepackage{lastpage}
\jmlrheading{22}{2021}{1-\pageref{LastPage}}{4/20; Revised
12/20}{4/21}{20-376}{Yasuhiro Fujita, Prabhat Nagarajan, Toshiki Kataoka, and Takahiro Ishikawa}
\ShortHeadings{ChainerRL: A Deep Reinforcement Learning Library}{Fujita, Nagarajan, Kataoka, and Ishikawa}

\firstpageno{1}

\begin{document}


\title{ChainerRL: A Deep Reinforcement Learning Library}

\author{\name Yasuhiro Fujita \email fujita@preferred.jp \\
       \name  Prabhat Nagarajan \email prabhat@preferred.jp \\
       \name  Toshiki Kataoka \email kataoka@preferred.jp \\
       Preferred Networks\\
       Tokyo, Japan
       \AND
       \name  Takahiro Ishikawa \email sykwer@g.ecc.u-tokyo.ac.jp \\
       The University of Tokyo \\
       Tokyo, Japan
       }

\editor{Andreas Mueller}

\maketitle

\begin{abstract}%
In this paper, we introduce \texttt{ChainerRL}, an open-source deep reinforcement learning (DRL) library built using Python and the \texttt{Chainer} deep learning framework.
\texttt{ChainerRL} implements a comprehensive set of DRL algorithms and techniques drawn from state-of-the-art research in the field. To foster reproducible research, and for instructional purposes, \crl{} provides scripts that closely replicate the original papers' experimental settings and reproduce published benchmark results for several algorithms.
Lastly, \texttt{ChainerRL} offers a visualization tool that enables the qualitative inspection of trained agents.
The \texttt{ChainerRL} source code can be found on GitHub: \url{https://github.com/chainer/chainerrl}.
\end{abstract}


\section{Introduction}
Since its resurgence in 2013~\citep{dqn2013}, deep reinforcement learning (DRL) has undergone tremendous progress, and has enabled significant advances in numerous complex sequential decision-making problems \citep{dqn, alphazero, Levine2016, Kalashnikov2018}. The machine learning community has witnessed a growing body of literature on DRL algorithms \citep{deeprlmatters}. However, coinciding with this rapid growth has been a growing concern about the state of reproducibility in DRL~\citep{deeprlmatters}. The growing body of algorithms and increased reproducibility concerns beget the need for comprehensive libraries, tools, and implementations that can aid RL-based research and development.

Many libraries aim to address these challenges in different ways.
\texttt{rllab}~\citep{rllab} and its successor, \texttt{garage}, provide systematic benchmarking of continuous-action algorithms on their own benchmark environments.
\texttt{Dopamine}~\citep{dopamine} primarily focuses on DQN and its extensions for discrete-action environments.
\texttt{rlpyt}~\citep{rlpyt} supports both discrete and continuous-action algorithms from the three classes: policy gradient (with V-functions), deep Q-learning, and policy gradient with Q-functions.
Other libraries also support diverse sets of algorithms~\citep{baselines, coach, stablebaselines, rayrllib}.
\texttt{catalyst.RL}~\citep{catalyst_rl} aims to address reproducibility issues in RL via deterministic evaluations and by tracking code changes for continuous-action algorithms.

In this paper, we introduce \crl{}, an open-source Python DRL library supporting both CPU and GPU training, built off of the \texttt{Chainer}~\citep{chainer} deep learning framework. \crl{} offers a comprehensive set of algorithms and abstractions, a set of ``reproducibility scripts'' that replicate research papers, and a companion visualizer to inspect agents.

\section{Design of ChainerRL} \label{overview}
\begin{figure}
    \centering
    \includegraphics[width=0.95\linewidth]{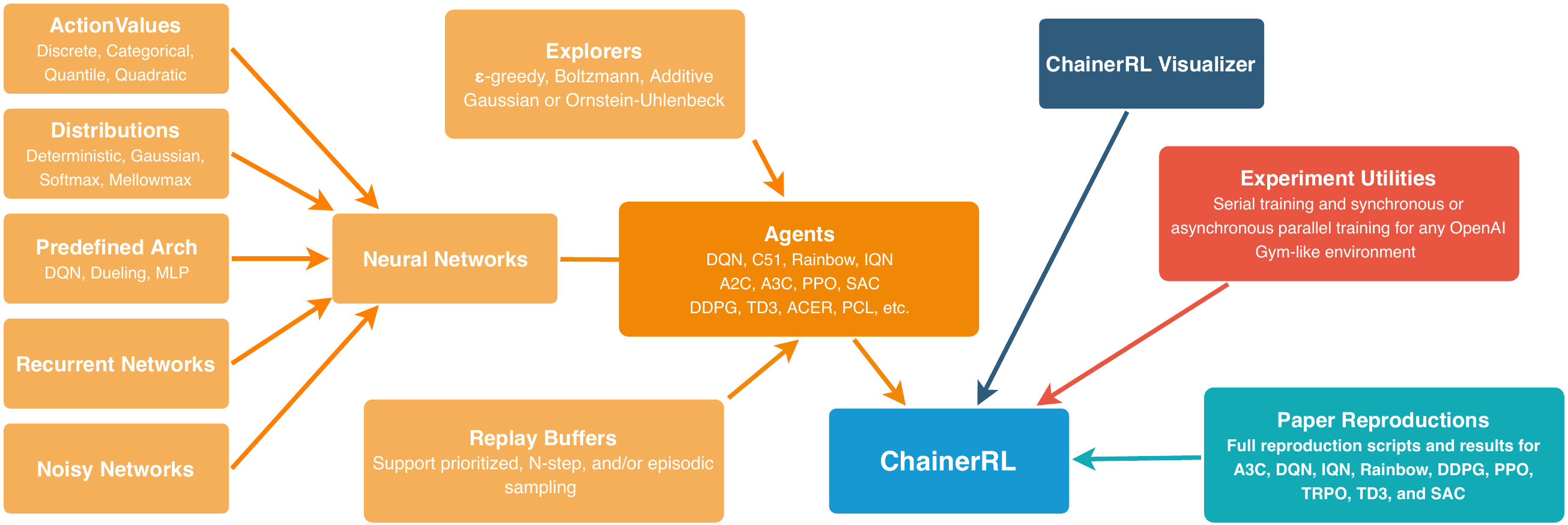}
    \caption{A depiction of \crl{}.
    Using \crl{}'s building blocks, DRL algorithms, called agents, are written by implementing the \texttt{Agent} interface. Agents can be trained with the experiment utilities and inspected with the \crlv{}.
    }
    \label{fig:crl}
\end{figure}

In this section, we describe \crl's design, as in Figure \ref{fig:crl}.
\subsection{Agents}
In \crl{}, each DRL algorithm is written as a class that implements the \texttt{Agent} interface.
The \texttt{Agent} interface provides a mechanism through which an agent interacts with an environment, e.g., through an abstract method \texttt{Agent.act\_and\_train(obs, reward, done)} that takes as input the current observation, the previous step's immediate reward, and a flag for episode termination, and returns the agent's action to execute in the environment.
By implementing such methods, both the update rule and the action-selection procedure are specified for an algorithm.

An agent's internals consist of any model parameters needed for decision-making and model updating.
\crl{} includes several built-in agents that implement key algorithms including the DQN~\citep{dqn} family of algorithms, as well as several policy gradient and actor-critic algorithms.\footnote{\crl{}'s algorithms include: DQN~\citep{dqn}, Double DQN~\citep{ddqn}, Categorical DQN~\citep{distributionaldqn}, Rainbow~\citep{rainbow}, Implicit Quantile Networks (IQN)~\citep{Dabney2018}, Off-policy SARSA, (Persistent) Advantage Learning~\citep{Bellemare2016c}, (Asynchronous) Advantage Actor-Critic (A2C~\citep{a2c}, A3C~\citep{a3c}), Actor-Critic with Experience Replay (ACER)~\citep{acer}, Deep Deterministic Policy Gradients (DDPG)~\citep{ddpg}, Twin-delayed double DDPG (TD3)~\citep{Fujimoto2018a}, Proximal Policy Optimization (PPO)~\citep{Schulman2017b}, REINFORCE~\citep{Williams1992}, Trust Region Policy Optimization (TRPO)~\citep{trpo}, and Soft Actor-Critic (SAC)~\citep{sac}.
}

\subsection{Experiments}
While users can directly interact with agents, \crl{} provides an \texttt{experiments} module that manages agent-environment interactions as well as training/evaluation schedules.
This module supports any environment that is compatible with \texttt{OpenAI Gym}'s \texttt{Env}~\citep{gym}.
An experiment takes as input an agent and an environment, queries the agent for actions, executes them in the environment, and feeds the agent the rewards for training updates.
Moreover, an experiment can periodically perform evaluations and collect evaluation statistics. Through the \texttt{experiments} module, \crl{} supports batch or asynchronous training, enabling agents to act, train, and evaluate synchronously or asynchronously in several environments in parallel. A full list of synchronous and asynchronous agents is provided in the appendix.

\subsection{Developing a New Agent}
The \texttt{Agent} interface is defined very abstractly and flexibly so that users can easily implement new algorithms while leveraging the \texttt{experiments} utility and parallel training infrastructure. To develop a new agent, we first create a class that inherits \texttt{Agent}.
Next, the learning update rules and the agent's action-selection mechanisms are implemented using \crl{}'s provided building blocks (see Section \ref{features}).
Once an agent is created, the agent and a \texttt{Gym}-like environment can be given to the \texttt{experiments} module to easily train and evaluate the agent within the specified environment.

\subsection{Agent Building Blocks} \label{features}
\crl{} offers a set of reusable components for building new agents, including \crl{}'s built-in agents.
Though not comprehensive, we highlight here some of the building blocks that demonstrate the flexibility and reusability of \crl{}.
\begin{description}
\itemsep0em 
\item[Explorers]
For building action-selection mechanisms during training, \crl{} has built-in explorers including $\epsilon$-greedy, Boltzmann exploration, additive Gaussian noise, and additive Ornstein-Uhlenbeck noise~\citep{ddpg}.
\item[Replay buffers]
Replay buffers~\citep{Lin1992, dqn} have become standard tools in off-policy DRL. \crl{} supports traditional uniform-sampling replay buffers, episodic buffers for sampling past (sub-)episodes for recurrent models, and prioritized buffers that prioritize sampled transitions~\citep{prioritizeddqn}. \crl{} also supports sampling $N$ steps of transitions, for algorithms based on $N$-step returns. 
\item[Neural networks]
While \crl{} supports any \texttt{Chainer} model, it has several pre-defined architectures, including DQN architectures, dueling network architectures~\citep{duelingdqn}, noisy networks~\citep{noisynetworks}, and multi-layer perceptrons.
Recurrent models are supported for many algorithms, including DQN and IQN.
\item[Distributions]
\texttt{Distribution}s are parameterized objects for modeling action distributions. 
Network models that return \texttt{Distribution} objects are considered policies.
Supported policies include Gaussian, Softmax, Mellowmax~\citep{Asadi2016a}, and deterministic policies.
\item[Action values]
Similar to \texttt{Distribution}s, \texttt{ActionValue}s parameterizing the values of actions are used as outputs of neural networks to model Q-functions.
Supported Q-functions include the standard discrete-action Q-function typical of DQN as well as categorical~\citep{distributionaldqn} and quantile~\citep{Dabney2018} Q-functions for distributional RL.
For continuous action spaces, quadratic Q-functions called Normalized Advantage Functions (NAFs)~\citep{Gu2016b} are also supported.
\end{description}

By combining these agent building blocks, users can easily construct complex agents such as Rainbow~\citep{rainbow}, which combines six features into a single agent. This ability is highlighted in Appendix \ref{pseudocode}, which provides a pseudocode construction of a Rainbow agent and trains it in multiple parallel environments in just a few lines.

\subsection{Visualization} \label{visualization}
\crl{} is accompanied by the \crlv{}, which takes as input an environment and an agent, and enables users to easily inspect agents from a browser UI.
With the visualizer, one can visualize the portions of the pixel input that the agent is attending to as a saliency map~\citep{visualizingatari}.
Additionally, users can either manually step through the episode or view full rollouts of agents. Moreover, the visualizer depicts the probabilities with which the agent will perform specific actions.
If the agent learns Q-values or a distribution of Q-values, the predicted Q-value or Q-value distribution for each action can be displayed. Figure \ref{fig:viz1} in Appendix \ref{appendix_visualizer} depicts some of these features.

\section{Reproducibility} \label{reproducibility}
Many DRL libraries offer implementations of algorithms but often deviate from the original paper's implementation details.
We provide a set of ``reproducibility scripts'', which are compact examples (i.e., single files) of paper implementations written with \crl{} that match, as closely as possible, the original paper's (or in some cases, another published paper's) implementation and evaluation details.
\crl{} currently has ``reproducibility scripts'' for DQN, IQN, Rainbow, A3C, DDPG, TRPO, PPO, TD3, and SAC. For each of these algorithms and domains, we have released pretrained models for every domain, totaling hundreds of models. Moreover, for each script, we provide full tables of our scores and compare them against scores reported in the literature (Tables \ref{atariresults} and \ref{mujocoresults} in Appendix \ref{appendix_reproducibility}).

\section{Conclusion} \label{conclusion}
This paper introduced \crl{} and the \crlv{}. \crl{}'s comprehensive suite of algorithms, flexible APIs, visualization tools, and faithful reproductions can accelerate the research and application of DRL algorithms.
While \crl{} targets Chainer users, we have developed an analogous library,  \texttt{PFRL}, for PyTorch users.\footnote{The \texttt{PFRL} code is located at \url{https://github.com/pfnet/pfrl}.}


\acks{We thank Avinash Ummadisingu, Mario Ynocente Castro, Keisuke Nakata, Lester James V. Miranda, and all the open source contributors for their contributions to the development of ChainerRL.
We thank Kohei Hayashi and Jason Naradowsky for useful comments on how to improve the paper.
We thank the many authors who fielded our questions when reproducing their papers, especially George Ostrovski.}



\newpage
\appendix

\section{Agents}
\crl{} implements several kinds of agents, supporting discrete-action agents, continuous-action agents, recurrent agents, batch agents, and asynchronous agents.
Asynchronous training, where an agent interacts with multiple environments asynchronously with a single set of model parameters, is supported for A3C, ACER~\citep{acer}, N-step Q-learning, and Path Consistency Learning (PCL).
To train an asynchronous agent, one can simply initialize an asynchronous agent and train it using \texttt{experiments.train\_agent\_async}.
Batch training refers to synchronous parallel training, where a single agent interacts with multiple environments synchronously in parallel, and is supported for all algorithms for which asynchronous training is not supported. 
In \crl{}, users can easily perform batch training of agents by initializing an agent and using \texttt{experiments.train\_agent\_batch\_with\_evaluation}.
Many algorithms require additional infrastructure to support recurrent training, e.g., by storing and managing the recurrent state, and managing sequences of observations as opposed to individual observations. 
\crl{} abstracts these difficulties away from the user, making it simple to employ recurrent architectures for the majority of algorithms.
Note that most of the algorithms implemented in \crl{} do not have support for recurrence or batch training in their original published form.
In \crl{}, we have added this additional support for most algorithms, as summarized in Table~\ref{alglisttable}.

\begin{table*}[ht] \centering
\begin{small}
\resizebox{\columnwidth}{!}{%
\begin{tabular}{@{}l|c|c|c|c|c@{}}\toprule
  \textbf{Algorithm} \hfill & \textbf{Discrete Action} & \textbf{Continuous Action} & \textbf{Recurrent Model} & \textbf{Batch Training} & \textbf{CPU Async Training}\\ \midrule

\textbf{DQN} (Double DQN, SARSA, etc.) & \cmark & \cmark (NAF) &  \cmark & \cmark & \xmark \\
\textbf{Categorical DQN} & \cmark & \xmark  &  \cmark & \cmark & \xmark \\
\textbf{Rainbow} & \cmark & \xmark &  \cmark & \cmark & \xmark \\
\textbf{IQN} (and Double IQN) & \cmark & \xmark &  \cmark & \cmark & \xmark \\
\textbf{A3C} & \cmark & \cmark &  \cmark & \cmark  (A2C) & \cmark \\
\textbf{ACER} & \cmark & \cmark &  \cmark & \xmark & \cmark \\
\textbf{NSQ (N-step Q-learning)} & \cmark & \cmark (NAF) &  \cmark & \xmark & \cmark \\
\textbf{PCL (Path Consistency Learning)} & \cmark & \cmark &  \cmark & \xmark & \cmark \\
\textbf{DDPG} & \xmark & \cmark &  \cmark & \cmark & \xmark \\
\textbf{PPO} & \cmark & \cmark &  \cmark & \cmark & \xmark \\
\textbf{TRPO} & \cmark & \cmark &  \cmark & \cmark & \xmark \\
\textbf{TD3} & \xmark & \cmark &  \xmark & \cmark & \xmark \\
\textbf{SAC} & \xmark & \cmark &  \xmark & \cmark & \xmark \\
\midrule
\bottomrule
\end{tabular}
}
\end{small}
\caption{Summarized list of \crl{} algorithms and their additional supported features.}
\label{alglisttable}
\end{table*}

\section{Reproducibility Results} \label{appendix_reproducibility}
For each of our reproducibility scripts, we provide the training times of the script (in our repository), full tables of our achieved scores, and comparisons of these scores against those reported in the literature.
Though \crl{} has high-quality implementations of dozens of algorithms, we currently have created such ``reproducibility scripts'' for 9 algorithms.
In the Atari benchmark~\citep{ale}, we have successfully reproduced DQN, IQN, Rainbow, and A3C.
For the OpenAI Gym Mujoco benchmark tasks, we have successfully reproduced DDPG, TRPO, PPO, TD3, and SAC.

The reproducibility scripts emphasize correctly reproducing evaluation protocols, which are particularly relevant when evaluating Atari agents. 
Unfortunately, evaluation protocols tend to vary across papers, and consequently results are often inconsistently reported across the literature~\citep{revisitingale}, significantly impacting results. 
The critical details of standard Atari evaluation protocols are as follows:
\begin{description}
\item[Evaluation frequency] The frequency (in timesteps) at which the evaluation phase occurs.
\item[Evaluation phase length] The number of timesteps in the offline evaluation.
\item[Evaluation episode length] The maximum duration of an evaluation episode.
\item[Evaluation policy] The policy to follow during an evaluation episode.
\item[Reporting protocol] Each intermediate evaluation phase outputs some score, representing the mean score of all evaluation episodes during that evaluation phase. Papers typically report scores according to one of the following reporting protocols: 
    \begin{enumerate}
        \item \textit{best-eval}: Papers using the \textit{best-eval} protocol report the highest mean score across all intermediate evaluation phases.
        \item \textit{re-eval}: Papers using the \textit{re-eval} protocol report the score of a re-evaluation of the network parameters that produced the \textit{best-eval}.
    \end{enumerate}
\end{description}

During a typical Atari agent's 50 million timesteps of training, it is evaluated periodically in an offline evaluation phase for a specified number of timesteps before resuming training. 
Since most papers report final results using the best model as determined by these periodic evaluation phases, the frequency of evaluation is key, as it provides the author of a paper with more models to select from when reporting final results.
The length of the evaluation phase is important, because shorter evaluation phases have higher variance in performance and longer evaluation phases have less variance in performance. 
Again, since these intermediate evaluations are used in some way when reporting final performance, the length of the evaluation phase is important when reproducing results.
The length of the evaluation episodes can impact performance, as permitting the agent to have longer episodes may allow it to accrue more points.
Oftentimes, since the agent performs some form of exploratory policy during training, the agent sometimes changes policies specifically for evaluations. 
Each of the listed details, especially the reporting protocols, can significantly influence the results, and thus are critical details to hold consistent for a fair comparison between algorithms.

Table~\ref{atariresults} lists the results obtained by \crl{}'s reproducibility scripts for DQN, IQN, Rainbow, and A3C on the Atari benchmark, with comparisons against a published result.
Table~\ref{evalprotocolatari} depicts the evaluation protocol used for each algorithm, with a citation of the source paper whose results we compare against. Note that the results for the A3C~\citep{a3c} algorithm do not come from the original A3C paper, but from another~\citep{noisynetworks}.
For continuous-action algorithms, the results on \texttt{OpenAI Gym} MuJoCo tasks for DDPG~\citep{ddpg}, TRPO~\citep{trpo}, PPO~\citep{Schulman2017b}, TD3~\citep{Fujimoto2018a}, and SAC~\citep{sac} are reported in Table~\ref{mujocoresults}. For all algorithms and environments listed in tables~\ref{atariresults} and~\ref{mujocoresults}, we have released models trained through our reproducibility scripts, which researchers can use.

The reproducibility scripts are produced through a combination of reading released source code and studying published hyperparameters, implementation details, and evaluation protocols. We also have extensive email correspondences with authors to clarify ambiguities, omitted details, or inconsistencies that may exist in papers.

As seen in both the Atari and MuJoCo reproducibility results, sometimes a reproduction effort cannot be directly compared against the original paper's reported results.
For example, the reported scores in the original paper introducing the A3C algorithm~\citep{a3c} utilize demonstrations that are not publicly available, making it impossible to accurately compare a re-implementation's scores to the original paper.
In such scenarios, we seek out high-quality published research~\citep{noisynetworks, deeprlmatters, Fujimoto2018a} from which faithful reproductions are indeed possible, and compare against these. 

\clearpage
\begin{table}[!t] \centering
\resizebox{\columnwidth}{!}{%
\begin{tabular}{@{}l|cc|cc|cc|cc@{}}\toprule
& \multicolumn{2}{c|}{\textbf{DQN}} & \multicolumn{2}{c|}{\textbf{IQN}} & \multicolumn{2}{c|}{\textbf{Rainbow}} & \multicolumn{2}{c}{\textbf{A3C}} \\
\midrule
\textbf{Game} \hfill & \textbf{CRL} & \textbf{Published}& \textbf{CRL} & \textbf{Published}& \textbf{CRL} & \textbf{Published}& \textbf{CRL} & \textbf{Published} \\ \midrule
\textsc{Air Raid} & 6450.5  $\pm$ 5.9e+2  &  - & 9933.5  $\pm$ 4.9e+2  &  - & 6754.3  $\pm$ 2.4e+2  &  - & 3923.8  $\pm$ 1.5e+2  &  - \\
\textsc{Alien} & 1713.1 $\pm$ 2.3e+2  & \textbf{3069} & \textbf{12049.2} $\pm$ 8.9e+2  & 7022 & \textbf{11255.4} $\pm$ 1.6e+3  & 9491.7 & 2005.4 $\pm$ 4.3e+2  & \textbf{2027} \\
\textsc{Amidar} & \textbf{986.7} $\pm$ 1.0e+2  & 739.5 & 2602.9 $\pm$ 3.9e+2  & \textbf{2946} & 3302.3 $\pm$ 7.2e+2  & \textbf{5131.2} & 869.7 $\pm$ 7.7e+1  & \textbf{904} \\
\textsc{Assault} & 3317.2 $\pm$ 7.3e+2  & \textbf{3359} & 24315.8 $\pm$ 9.6e+2  & \textbf{29091} & \textbf{17040.6} $\pm$ 2.0e+3  & 14198.5 & \textbf{6832.6} $\pm$ 2.e+3  & 2879 \\
\textsc{Asterix} & 5936.7 $\pm$ 7.3e+2  & \textbf{6012} & \textbf{484527.4} $\pm$ 7.4e+4  & 342016 & \textbf{440208.0} $\pm$ 9.e+4  & 428200.3 & \textbf{9363.0} $\pm$ 2.8e+3  & 6822 \\
\textsc{Asteroids} & 1584.5 $\pm$ 1.6e+2  & \textbf{1629} & \textbf{3806.2} $\pm$ 1.5e+2  & 2898 & \textbf{3274.9} $\pm$ 8.4e+2  & 2712.8 & \textbf{2775.6} $\pm$ 3.3e+2  & 2544 \\
\textsc{Atlantis} & \textbf{96456.0} $\pm$ 6.5e+3  & 85641 & 937491.7 $\pm$ 1.6e+4  & \textbf{978200} & \textbf{895215.8} $\pm$ 1.3e+4  & 826659.5 & \textbf{836040.0} $\pm$ 4.7e+4  & 422700 \\
\textsc{Bank Heist} & \textbf{645.0} $\pm$ 4.7e+1  & 429.7 & 1333.2 $\pm$ 2.3e+1  & \textbf{1416} & \textbf{1655.1} $\pm$ 1.0e+2  & 1358.0 & \textbf{1321.6} $\pm$ 6.6e+0  & 1296 \\
\textsc{Battle Zone} & 5313.3 $\pm$ 2.9e+3  & \textbf{26300} & \textbf{67834.0} $\pm$ 5.1e+3  & 42244 & \textbf{87015.0} $\pm$ 1.3e+4  & 62010.0 & 7998.0 $\pm$ 2.6e+3  & \textbf{16411} \\
\textsc{Beam Rider} & \textbf{7042.9} $\pm$ 5.2e+2  & 6846 & 40077.2 $\pm$ 4.1e+3  & \textbf{42776} & \textbf{26672.1} $\pm$ 8.3e+3  & 16850.2 & 9044.4 $\pm$ 4.7e+2  & \textbf{9214} \\
\textsc{Berzerk} & 707.2  $\pm$ 1.7e+2  &  - & \textbf{92830.5} $\pm$ 1.6e+5  & 1053 & \textbf{17043.4} $\pm$ 1.2e+4  & 2545.6 & \textbf{1166.8} $\pm$ 3.8e+2  & 1022 \\
\textsc{Bowling} & \textbf{52.3} $\pm$ 1.2e+1  & 42.4 & 85.8 $\pm$ 6.1e+0  & \textbf{86.5} & \textbf{55.7} $\pm$ 1.5e+1  & 30.0 & 31.3 $\pm$ 2.4e-1  & \textbf{37} \\
\textsc{Boxing} & \textbf{89.6} $\pm$ 3.1e+0  & 71.8 & \textbf{99.9} $\pm$ 2.1e-2  & 99.8 & \textbf{99.8} $\pm$ 1.3e-1  & 99.6 & \textbf{96.0} $\pm$ 1.9e+0  & 91 \\
\textsc{Breakout} & 364.9 $\pm$ 3.4e+1  & \textbf{401.2} & 665.2 $\pm$ 1.1e+1  & \textbf{734} & 353.0 $\pm$ 1.1e+1  & \textbf{417.5} & \textbf{569.9} $\pm$ 1.9e+1  & 496 \\
\textsc{Carnival} & 5222.0  $\pm$ 2.9e+2  &  - & 5478.7  $\pm$ 4.6e+2  &  - & 4762.8  $\pm$ 6.6e+2  &  - & 4643.3  $\pm$ 1.2e+3  &  - \\
\textsc{Centipede} & 5112.6 $\pm$ 6.9e+2  & \textbf{8309} & 10576.6 $\pm$ 1.7e+3  & \textbf{11561} & \textbf{8220.1} $\pm$ 4.6e+2  & 8167.3 & \textbf{5352.4} $\pm$ 3.3e+2  & 5350 \\
\textsc{Chopper Command} & 6170.0 $\pm$ 1.6e+3  & \textbf{6687} & \textbf{39400.9} $\pm$ 7.4e+3  & 16836 & \textbf{103942.2} $\pm$ 1.7e+5  & 16654.0 & \textbf{6997.1} $\pm$ 4.5e+3  & 5285 \\
\textsc{Crazy Climber} & 108472.7 $\pm$ 1.5e+3  & \textbf{114103} & 178080.2 $\pm$ 3.0e+3  & \textbf{179082} & \textbf{174438.8} $\pm$ 1.8e+4  & 168788.5 & 121146.1 $\pm$ 2.6e+3  & \textbf{134783} \\
\textsc{Demon Attack} & 9044.3 $\pm$ 1.8e+3  & \textbf{9711} & \textbf{135497.1} $\pm$ 1.5e+3  & 128580 & 101076.9 $\pm$ 1.1e+4  & \textbf{111185.2} & \textbf{111339.2} $\pm$ 6.3e+3  & 37085 \\
\textsc{Double Dunk} & \textbf{-9.7} $\pm$ 1.8e+0  & -18.1 & 5.6 $\pm$ 1.4e+1  & 5.6& -1.0 $\pm$ 7.9e-1  & \textbf{-0.3} & 1.5 $\pm$ 3.5e-1  & \textbf{3} \\
\textsc{Enduro} & 298.2 $\pm$ 5.4e+0  & \textbf{301.8} & \textbf{2363.6} $\pm$ 3.3e+0  & 2359 & \textbf{2278.6} $\pm$ 4.1e+0  & 2125.9 & 0.0 $\pm$ 0.e+0  & 0\\
\textsc{Fishing Derby} & \textbf{11.6} $\pm$ 7.6e+0  & -0.8 & \textbf{38.8} $\pm$ 4.3e+0  & 33.8 & \textbf{44.6} $\pm$ 5.1e+0  & 31.3 & \textbf{38.7} $\pm$ 1.6e+0  & -7 \\
\textsc{Freeway} & 8.1 $\pm$ 1.3e+1  & \textbf{30.3} & 34.0 $\pm$ 0.e+0  & 34.0& 33.6 $\pm$ 4.6e-1  & \textbf{34.0} & 0.0 $\pm$ 7.3e-3  & 0\\
\textsc{Frostbite} & \textbf{1093.9} $\pm$ 5.5e+2  & 328.3 & \textbf{8196.1} $\pm$ 1.5e+3  & 4342 & \textbf{10071.6} $\pm$ 8.6e+2  & 9590.5 & \textbf{288.2} $\pm$ 2.9e+1  & 288 \\
\textsc{Gopher} & 8370.0 $\pm$ 1.1e+3  & \textbf{8520} & 117115.0 $\pm$ 2.8e+3  & \textbf{118365} & \textbf{82497.8} $\pm$ 5.6e+3  & 70354.6 & \textbf{9251.0} $\pm$ 1.8e+3  & 7992 \\
\textsc{Gravitar} & \textbf{445.7} $\pm$ 5.e+1  & 306.7 & \textbf{1006.7} $\pm$ 2.5e+1  & 911 & \textbf{1605.6} $\pm$ 1.9e+2  & 1419.3 & 244.5 $\pm$ 4.4e+0  & \textbf{379} \\
\textsc{Hero} & \textbf{20538.7} $\pm$ 2.0e+3  & 19950 & \textbf{28429.4} $\pm$ 2.4e+3  & 28386 & 27830.8 $\pm$ 1.3e+4  & \textbf{55887.4} & \textbf{36599.2} $\pm$ 3.5e+2  & 30791 \\
\textsc{Ice Hockey} & -2.4 $\pm$ 4.3e-1  & \textbf{-1.6} & 0.1 $\pm$ 2.0e+0  & \textbf{0.2} & \textbf{5.7} $\pm$ 5.4e-1  & 1.1 & -4.5 $\pm$ 1.9e-1  & \textbf{-2} \\
\textsc{Jamesbond} & \textbf{851.7} $\pm$ 2.3e+2  & 576.7 & 26033.6 $\pm$ 3.8e+3  & \textbf{35108} & 24997.6  $\pm$ 5.6e+3  &  - & 376.9 $\pm$ 2.6e+1  & \textbf{509} \\
\textsc{Journey Escape} & -1894.0  $\pm$ 5.8e+2  &  - & -632.9  $\pm$ 9.7e+1  &  - & -429.2  $\pm$ 4.4e+2  &  - & -989.2  $\pm$ 4.2e+1  &  - \\
\textsc{Kangaroo} & \textbf{8831.3} $\pm$ 6.8e+2  & 6740 & \textbf{15876.3} $\pm$ 6.4e+2  & 15487 & 11038.8 $\pm$ 5.8e+3  & \textbf{14637.5} & 252.0 $\pm$ 1.2e+2  & \textbf{1166} \\
\textsc{Krull} & \textbf{6215.0} $\pm$ 2.3e+3  & 3805 & 9741.8 $\pm$ 1.2e+2  & \textbf{10707} & 8237.9 $\pm$ 2.2e+2  & \textbf{8741.5} & 8949.3 $\pm$ 8.5e+2  & \textbf{9422} \\
\textsc{Kung Fu Master} & \textbf{27616.7} $\pm$ 1.3e+3  & 23270 & \textbf{87648.3} $\pm$ 1.1e+4  & 73512 & 33628.2 $\pm$ 9.5e+3  & \textbf{52181.0} & \textbf{39676.3} $\pm$ 2.4e+3  & 37422 \\
\textsc{Montezuma Revenge} & 0.0 $\pm$ 0.e+0  & 0.0& \textbf{0.4} $\pm$ 6.8e-1  & 0.0 & 16.2 $\pm$ 2.2e+1  & \textbf{384.0} & 2.8 $\pm$ 6.3e-1  & \textbf{14} \\
\textsc{Ms Pacman} & \textbf{2526.6} $\pm$ 1.e+2  & 2311 & 5559.7 $\pm$ 4.5e+2  & \textbf{6349} & \textbf{5780.6} $\pm$ 4.6e+2  & 5380.4 & \textbf{2552.9} $\pm$ 1.9e+2  & 2436 \\
\textsc{Name This Game} & 7046.5 $\pm$ 2.0e+2  & \textbf{7257} & \textbf{23037.2} $\pm$ 2.e+2  & 22682 & \textbf{14236.4} $\pm$ 8.5e+2  & 13136.0 & \textbf{8646.0} $\pm$ 3.e+3  & 7168 \\
\textsc{Phoenix} & 7054.4  $\pm$ 1.9e+3  &  - & \textbf{125757.5} $\pm$ 3.6e+4  & 56599 & 84659.6 $\pm$ 1.4e+5  & \textbf{108528.6} & \textbf{38428.3} $\pm$ 3.1e+3  & 9476 \\
\textsc{Pitfall} & -28.3  $\pm$ 2.1e+1  &  - & 0.0 $\pm$ 0.e+0  & 0.0& -3.2 $\pm$ 2.9e+0  & \textbf{0.0} & -4.4 $\pm$ 2.9e+0  & \textbf{0} \\
\textsc{Pong} & \textbf{20.1} $\pm$ 4.0e-1  & 18.9 & 21.0 $\pm$ 0.e+0  & 21.0& \textbf{21.0} $\pm$ 6.4e-2  & 20.9 & \textbf{20.7} $\pm$ 3.9e-1  & 7 \\
\textsc{Pooyan} & 3118.7  $\pm$ 3.5e+2  &  - & 27222.4  $\pm$ 9.9e+3  &  - & 7772.7  $\pm$ 3.6e+2  &  - & 4237.9  $\pm$ 5.8e+1  &  - \\
\textsc{Private Eye} & 1538.3 $\pm$ 1.3e+3  & \textbf{1788} & \textbf{259.9} $\pm$ 1.0e+2  & 200 & 99.3 $\pm$ 5.8e-1  & \textbf{4234.0} & 449.0 $\pm$ 1.6e+2  & \textbf{3781} \\
\textsc{Qbert} & 10516.0 $\pm$ 2.6e+3  & \textbf{10596} & 25156.8 $\pm$ 5.3e+2  & \textbf{25750} & \textbf{41819.6} $\pm$ 1.9e+3  & 33817.5 & \textbf{18889.2} $\pm$ 7.6e+2  & 18586 \\
\textsc{Riverraid} & 7784.1 $\pm$ 6.8e+2  & \textbf{8316} & \textbf{21159.7} $\pm$ 8.0e+2  & 17765 & 26574.2  $\pm$ 1.8e+3  &  - & 12683.5  $\pm$ 5.3e+2  &  - \\
\textsc{Road Runner} & \textbf{37092.0} $\pm$ 3.e+3  & 18257 & \textbf{65571.3} $\pm$ 5.6e+3  & 57900 & \textbf{65579.3} $\pm$ 6.1e+3  & 62041.0 & 40660.6 $\pm$ 2.1e+3  & \textbf{45315} \\
\textsc{Robotank} & 47.4 $\pm$ 3.6e+0  & \textbf{51.6} & \textbf{77.0} $\pm$ 1.3e+0  & 62.5 & \textbf{75.6} $\pm$ 2.1e+0  & 61.4 & 3.1 $\pm$ 5.1e-2  & \textbf{6} \\
\textsc{Seaquest} & \textbf{6075.7} $\pm$ 2.3e+2  & 5286 & 26042.3 $\pm$ 3.9e+3  & \textbf{30140} & 3708.5 $\pm$ 1.7e+3  & \textbf{15898.9} & \textbf{1785.6} $\pm$ 4.1e+0  & 1744 \\
\textsc{Skiing} & -13030.2  $\pm$ 1.2e+3  &  - & -9333.6 $\pm$ 7.4e+1  & \textbf{-9289} & \textbf{-10270.9} $\pm$ 8.6e+2  & -12957.8 & -13094.2 $\pm$ 3.7e+3  & \textbf{-12972} \\
\textsc{Solaris} & 1565.1  $\pm$ 6.e+2  &  - & 7641.6 $\pm$ 8.2e+2  & \textbf{8007} & \textbf{8113.0} $\pm$ 1.2e+3  & 3560.3 & 3784.2 $\pm$ 3.5e+2  & \textbf{12380} \\
\textsc{Space Invaders} & 1583.2 $\pm$ 1.5e+2  & \textbf{1976} & \textbf{36952.7} $\pm$ 2.9e+4  & 28888 & 17902.6 $\pm$ 1.3e+4  & \textbf{18789.0} & \textbf{1568.9} $\pm$ 3.7e+2  & 1034 \\
\textsc{Star Gunner} & 56685.3 $\pm$ 1.0e+3  & \textbf{57997} & \textbf{182105.3} $\pm$ 1.9e+4  & 74677 & \textbf{188384.2} $\pm$ 2.3e+4  & 127029.0 & \textbf{60348.7} $\pm$ 2.6e+3  & 49156 \\
\textsc{Tennis} & -5.4 $\pm$ 7.6e+0  & \textbf{-2.5} & \textbf{23.7} $\pm$ 1.7e-1  & 23.6 & -0.0 $\pm$ 2.4e-2  & 0.0& -12.2 $\pm$ 4.3e+0  & \textbf{-6} \\
\textsc{Time Pilot} & 5738.7 $\pm$ 9.0e+2  & \textbf{5947} & \textbf{13173.7} $\pm$ 7.4e+2  & 12236 & \textbf{24385.2} $\pm$ 3.5e+3  & 12926.0 & 4506.6 $\pm$ 2.8e+2  & \textbf{10294} \\
\textsc{Tutankham} & 141.9 $\pm$ 5.1e+1  & \textbf{186.7} & \textbf{342.1} $\pm$ 8.2e+0  & 293 & \textbf{243.2} $\pm$ 2.9e+1  & 241.0 & \textbf{296.7} $\pm$ 1.8e+1  & 213 \\
\textsc{Up N Down} & \textbf{11821.5} $\pm$ 1.1e+3  & 8456 & 73997.8 $\pm$ 1.7e+4  & \textbf{88148} & 291785.9  $\pm$ 7.3e+3  &  - & \textbf{95014.6} $\pm$ 5.1e+4  & 89067 \\
\textsc{Venture} & \textbf{656.7} $\pm$ 5.5e+2  & 380.0 & 656.2 $\pm$ 6.4e+2  & \textbf{1318} & \textbf{1462.3} $\pm$ 3.4e+1  & 5.5 & 0.0 $\pm$ 0.e+0  & 0\\
\textsc{Video Pinball} & 9194.5 $\pm$ 6.3e+3  & \textbf{42684} & 664174.2 $\pm$ 1.1e+4  & \textbf{698045} & 477238.7 $\pm$ 2.6e+4  & \textbf{533936.5} & \textbf{377939.3} $\pm$ 1.8e+5  & 229402 \\
\textsc{Wizard Of Wor} & 1957.3 $\pm$ 2.7e+2  & \textbf{3393} & 23369.5 $\pm$ 5.4e+3  & \textbf{31190} & \textbf{20695.0} $\pm$ 9.e+2  & 17862.5 & 2518.7 $\pm$ 5.1e+2  & \textbf{8953} \\
\textsc{Yars Revenge} & 4397.3  $\pm$ 2.1e+3  &  - & \textbf{30510.0} $\pm$ 2.3e+2  & 28379 & 86609.9 $\pm$ 1.e+4  & \textbf{102557.0} & 19663.9 $\pm$ 6.6e+3  & \textbf{21596} \\
\textsc{Zaxxon} & \textbf{5698.7} $\pm$ 1.0e+3  & 4977 & 16668.5 $\pm$ 3.4e+3  & \textbf{21772} & \textbf{24107.5} $\pm$ 2.4e+3  & 22209.5 & 78.9 $\pm$ 6.8e+0  & \textbf{16544} \\
\midrule
\textbf{\# Higher scores} & 22 & 26 & 28 & 23 & 34 & 17 & 27 & 24 \\
\midrule
\textbf{\# Ties} & \multicolumn{2}{c|} 1 & \multicolumn{2}{c|} 4 & \multicolumn{2}{c|} 1 & \multicolumn{2}{c} 3 \\
\midrule
\textbf{\# Seeds} & 5 & 1 & 3 & 1 & 3 & 1 & 5 & 3  \\
\midrule
\bottomrule
\end{tabular}
}
\caption{The performance of \crl{} ($\pm$ standard deviation) against published results on Atari benchmarks.}
\label{atariresults}
\end{table}

\clearpage

\begin{table*}[t] \centering
\begin{small}
\begin{tabular}{@{}l|c|c|c|c@{}} \toprule
  & \textbf{DQN} & \textbf{IQN} & \textbf{Rainbow} & \textbf{A3C} \\
 \midrule
 \textbf{Eval Frequency} (timesteps) \hfill & 250K &  250K &  250K &  250K \\ 
 \textbf{Eval Phase} (timesteps) \hfill & 125K & 125K & 125K & 125K  \\ 
 \textbf{Eval Episode Length} (time) \hfill & 5 min & 30 min & 30 min & 30 min \\ 
 \textbf{Eval Episode Policy} \hfill & $\epsilon=0.05$ & $\epsilon=0.001$ & $\epsilon=0.0$ & N/A  \\ 
 \textbf{Reporting Protocol} \hfill & \textit{re-eval} & \textit{best-eval} & \textit{re-eval} &  \textit{best-eval} \\ 
\bottomrule
\end{tabular}
\end{small}
\caption{Evaluation protocols used for the Atari reproductions. The evaluation protocols of DQN, IQN, Rainbow, and A3C match the evaluation protocols used by \citet{dqn}, \citet{Dabney2018}, \citet{rainbow}, and \citet{noisynetworks}, respectively.
An evaluation episode policy with an $\epsilon$ indicates that the agent performs an $\epsilon$-greedy evaluation.
}
\label{evalprotocolatari}
\end{table*}

\begin{table*}[!t] \centering
\resizebox{\columnwidth}{!}{
\begin{tabular}{@{}l|cc|cc@{}}\toprule
                                   & \multicolumn{2}{c|}{\textbf{DDPG}~\citep{Fujimoto2018a}}     & \multicolumn{2}{c}{\textbf{TD3}~\citep{Fujimoto2018a}}                              \\
\midrule
\textbf{Environment} \hfill        & \textbf{CRL}& \textbf{Published} & \textbf{CRL}              & \textbf{Published}           \\ \midrule
\textsc{HalfCheetah-v2}            & \textbf{10325.45} & 8577.29            & \textbf{10248.51} $\pm$ 1063.48 & 9636.95 $\pm$ 859.065        \\
\textsc{Hopper-v2}                 & \textbf{3565.60}  & 1860.02            & \textbf{3662.85} $\pm$ 144.98   & 3564.07 $\pm$ 114.74         \\
\textsc{Walker2d-v2}               & \textbf{3594.26}  & 3098.11            & \textbf{4978.32} $\pm$ 517.44   & 4682.82 $\pm$ 539.64         \\
\textsc{Ant-v2}                    & \textbf{774.46}   & 888.77             & \textbf{4626.25} $\pm$ 1020.70  & 4372.44 $\pm$ 1000.33        \\
\textsc{Reacher-v2}                & \textbf{-2.92}    & -4.01              & \textbf{-2.55} $\pm$ 0.19       & -3.60 $\pm$ 0.56             \\
\textsc{InvertedPendulum-v2}       & 902.25            & \textbf{1000.00}   & 1000.00 $\pm$ 0.0               & 1000.00 $\pm$ 0.0            \\
\textsc{InvertedDoublePendulum-v2} & 7495.56           & \textbf{8369.95}   & 8435.33 $\pm$ 2771.39           & \textbf{9337.47} $\pm$ 14.96 \\
\bottomrule
\end{tabular}
}
\resizebox{\columnwidth}{!}{
\begin{tabular}{@{}l|cc|cc|cc@{}}\toprule
                                   & \multicolumn{2}{c|}{\textbf{TRPO}~\citep{deeprlmatters}}               & \multicolumn{2}{c|}{\textbf{PPO}~\citep{deeprlmatters}}                & \multicolumn{2}{c}{\textbf{SAC}~\citep{sac}}           \\
\midrule
\textbf{Environment} \hfill        & \textbf{CRL}      & \textbf{Published}     & \textbf{CRL}      & \textbf{Published}     & \textbf{CRL} & \textbf{Published}    \\ \midrule
\textsc{HalfCheetah-v2}            & \textbf{1474} $\pm$ 112 & 205  $\pm$ 256         & \textbf{2404} $\pm$ 185 & 2201 $\pm$ 323         & 14850.54           & \textasciitilde 15000 \\
\textsc{Hopper-v2}                 & \textbf{3056} $\pm$ 44  & 2828 $\pm$ 70          & 2719 $\pm$ 67           & \textbf{2790} $\pm$ 62 & 2911.89            & \textasciitilde 3300  \\
\textsc{Walker2d-v2}               & 3073 $\pm$ 59           & -                      & 2994 $\pm$ 113          & -                      & 5282.61            & \textasciitilde 5600  \\
\textsc{Ant-v2}                    & -                       & -                      & -                       & -                      & 5925.63            & \textasciitilde 5800  \\
\textsc{Swimmer-v2}                & 200 $\pm$ 25            & -                      & 111 $\pm$ 4             & -                      & -                  & -                     \\
\textsc{Humanoid-v2}               & -                       & -                      & -                       & -                      & 7772.08            & \textasciitilde 8000  \\
\bottomrule
\end{tabular}
}

\caption{
The performance of \crl{} against published baselines on \texttt{OpenAI Gym} MuJoCo benchmarks. For DDPG and TD3, each \texttt{ChainerRL} score represents the maximum evaluation score during 1M-step training, averaged over 10 trials with different random seeds, where each evaluation phase of ten episodes is run after every 5000 steps.
For PPO and TRPO, each \texttt{ChainerRL} score represents the final evaluation of 100 episodes after 2M-step training, averaged over 10 trials with different random seeds.
For SAC, each \texttt{ChainerRL} score reports the final evaluation of 10 episodes after training for 1M (Hopper-v2), 3M (HalfCheetah-v2, Walker2d-v2, and Ant-v2), or 10M (Humanoid-v2) steps, averaged over 10 trials with different random seeds.
Since the original paper ~\citep{sac} provides learning curves only, the published scores are approximated visually from the learning curve.
The sources of the published scores are cited with each algorithm.
We use the v2 environments, whereas some published papers evaluate on the now-deprecated v1 environments.
}
\label{mujocoresults}
\end{table*}

\section{Visualizer Images} \label{appendix_visualizer}

\begin{figure}[t]
    \centering
    \includegraphics[width=0.97\linewidth]{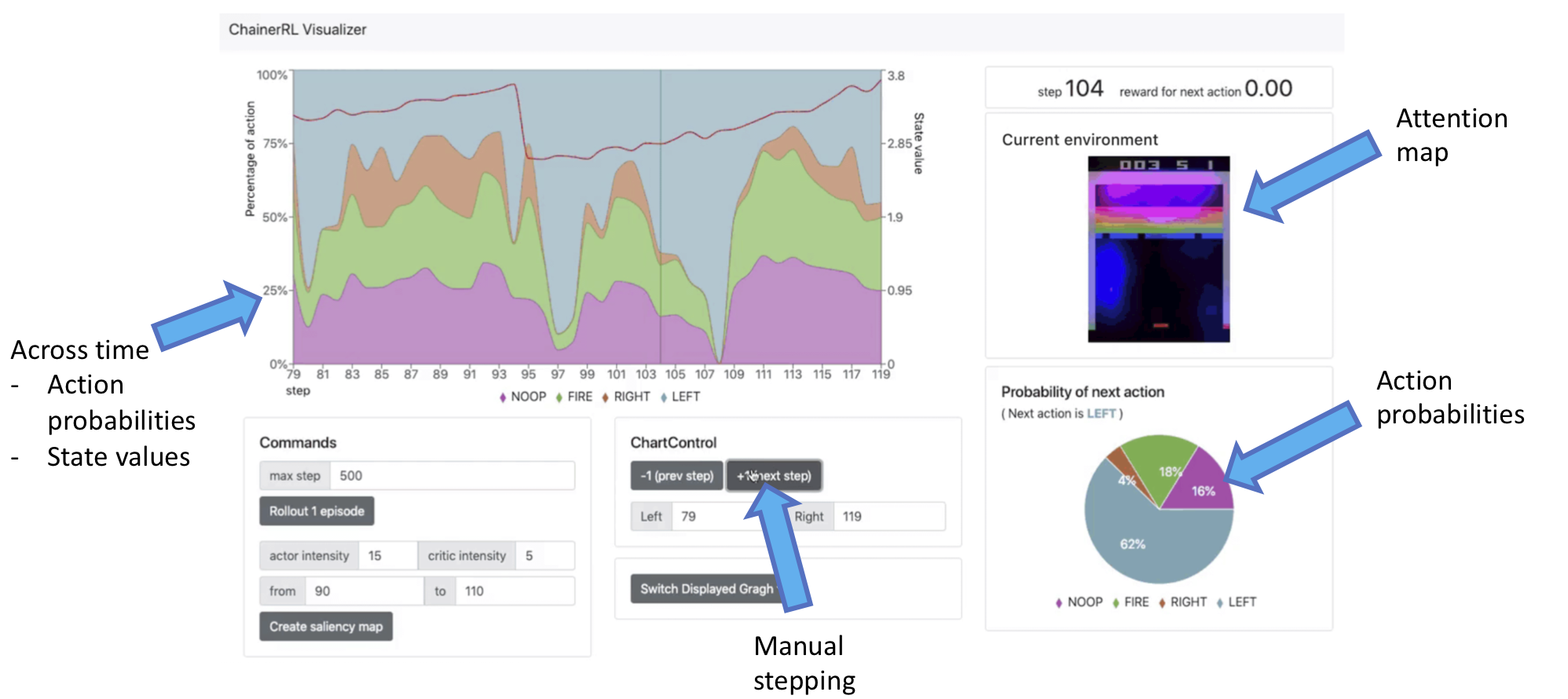}
    \includegraphics[width=0.97\linewidth]{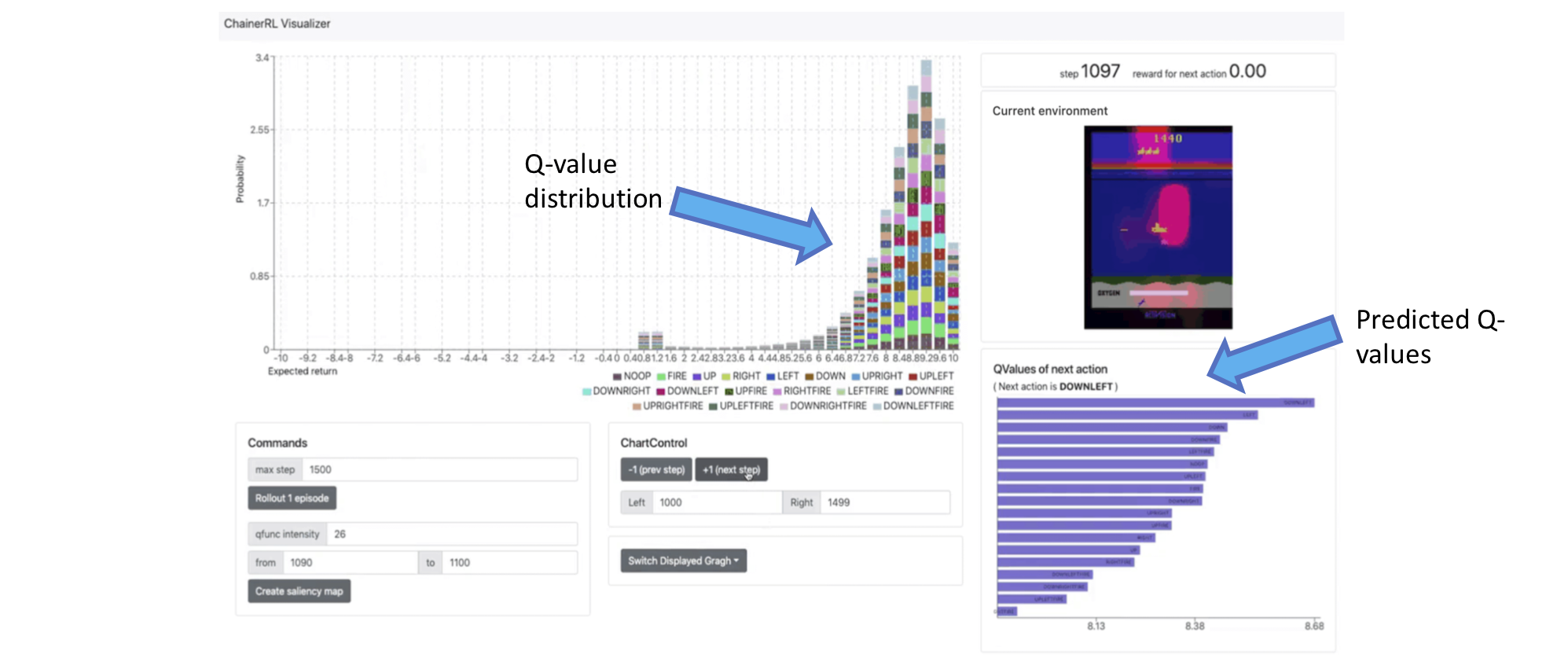}
    \caption{
The \crlv{}. With the \crlv{}, users can closely investigate an agent's behaviors within a browser window.
\emph{top}: Visualization of a trained A3C agent on \textsc{Breakout}. \emph{bottom}: Visualization of a C51 \citep{distributionaldqn} agent trained on \textsc{Seaquest}.}
    \label{fig:viz1}
\end{figure}

Figure~\ref{fig:viz1} depicts some of the key features of the \crlv{} for an actor-critic algorithm and a distributional value-based algorithm.
The top of the figure depicts a trained A3C agent in the Atari game \textsc{Breakout}.
With the visualizer, one can visualize the portions of the pixel input that the agent is attending to as a saliency map~\citep{visualizingatari}.
Additionally, users can perform careful, controlled investigations of agents by manually stepping through an episode, or can alternatively view rollouts of agents.
Since A3C is an actor-critic agent with a value function and a policy outputting a distribution over actions, we can view the probabilities with which the agent will perform a specific action, as well as the agent's predicted state values.
If the agent learns Q-values or a distribution of Q-values, the predicted Q-value or Q-value distribution for each action can be displayed, as shown in the bottom of Figure \ref{fig:viz1}.

\section{Pseudocode} \label{pseudocode}

The set of algorithms that can be developed by combining the agent building blocks of \crl{} is large.
One notable example is Rainbow~\citep{rainbow}, which combines double updating~\citep{ddqn}, prioritized replay~\citep{prioritizeddqn}, $N$-step learning, dueling architectures~\citep{duelingdqn}, and Categorical DQN~\citep{distributionaldqn} into a single agent.
The following pseudocode depicts the simplicity of creating and training a Rainbow agent with \crl{}. 

\begin{lstlisting}[language=Python]
import chainerrl as crl
import gym

q_func = crl.q_functions.DistributionalDuelingDQN(...)# dueling
crl.links.to_factorized_noisy(q_func) # noisy networks
# Prioritized Experience Replay Buffer with a 3-step reward
per = crl.replay_buffers.PrioritizedReplayBuffer(num_step_return=3,...)
# Create a rainbow agent
rainbow = crl.agents.CategoricalDoubleDQN(per, q_func,...)
num_envs = 5 # Train in five environments
env = crl.envs.MultiprocessVectorEnv(
    [gym.make("Breakout") for _ in range(num_envs)])

# Train the agent and collect evaluation statistics
crl.experiments.train_agent_batch_with_evaluation(rainbow, env, steps=...)
\end{lstlisting}

We first create a distributional dueling Q-function, and then in a single line, convert it to a noisy network.
We then initialize a prioritized replay buffer configured to use $N$-step rewards.
We pass this replay buffer to \crl{}'s built-in \texttt{CategoricalDoubleDQN} agent to produce a Rainbow agent.
Moreover, with \crl{}, users can easily specify the number of environments in which to train the Rainbow agent in synchronous parallel processes, and the \texttt{experiments} module will automatically manage the training loops, evaluation statistics, logging, and saving of the agent.


\end{document}